\documentclass{article}
\usepackage{spconf,amsmath,graphicx}
\usepackage{times}
\usepackage{epsfig}
\usepackage{amsmath}
\usepackage{amssymb}
\usepackage{mathtools}
\usepackage{multirow}
\usepackage{xcolor}
\usepackage{xurl}
\usepackage{makecell,rotating}
\usepackage[labelsep=period]{caption}
\newcommand\blfootnote[1]{%
	\begingroup
	\renewcommand\thefootnote{}\footnote{#1}%
	\addtocounter{footnote}{-1}%
	\endgroup
}

\usepackage{bbding}
\usepackage{amsmath}
\usepackage{amsfonts}
\usepackage{algorithm}
\usepackage{algorithmic}
\usepackage{caption}
\usepackage{subfigure}
\usepackage{graphicx}

\title{Learning to Locate Visual Answer in Video Corpus Using Question}
\name{Bin Li*$^\dag$, Yixuan Weng*$^\ddagger$, Bin Sun$^\dag$, Shutao Li$^{\dag\star}$}
\address{$^\dag$	College of Electrical and Information Engineering, Hunan University \\ $^\ddagger$ National Laboratory of Pattern Recognition Institute of Automation, Chinese Academy Sciences}

\begin{document}
	%
	
	\maketitle
	
	%
	%
	\begin{abstract}
We introduce a new task, named video corpus visual answer localization (VCVAL), which aims to locate the visual answer in a large collection of untrimmed instructional videos using a natural language question. This task requires a range of skills - the interaction between vision and language, video retrieval, passage comprehension, and visual answer localization. In this paper, we propose a cross-modal contrastive global-span (CCGS) method for the VCVAL, jointly training the video corpus retrieval and visual answer localization subtasks with the global-span matrix. We have reconstructed a dataset named MedVidCQA, on which the VCVAL task is benchmarked. Experimental results show that the proposed method outperforms other competitive methods both in the video corpus retrieval and visual answer localization subtasks.  Most importantly, we perform detailed analyses on extensive experiments, paving a new path for understanding the instructional videos, which ushers in further research\footnote{All the experimental datasets and codes are open-sourced on the website {https://github.com/WENGSYX/CCGS}.}.

	\end{abstract}
	\begin{keywords}
		Video corpus, visual answer localization
		\blfootnote{This work is supported by the National Natural Science Fund of China (62221002, 62171183) , the Hunan Provincial Natural Science Foundation of China (2022JJ20017), and partially sponsored by CAAI-Huawei MindSpore Open Fund.}
		\blfootnote{*: These authors contributed equally to this work.}
		\blfootnote{$^{\star}$: Corresponding author.}
	\end{keywords}
	\vspace{-0.3cm}
	\section{Introduction}
	In recent years, the popularity of the video platform has enriched people's lives 
	\cite{koupaee2018wikihow, drozd2018medical}. People can easily use natural language to query video, but when it comes to instructional or educational questions, more visual details are often needed to aid comprehension \cite{gupta2022dataset,li-etal-2022-vpai}. Thus, an increasing number of people expect to use the most intuitive video clips to answer directly when querying. Different from traditional video question-answering \cite{lei2018tvqa}, visual answer localization (VAL) is more efficient, which aims to provide a visual answer and has received extensive attention from researchers \cite{li2022towards, gupta2022overview}.
	\par
\begin{figure}[t]
		\centering
		\includegraphics[width=0.99\linewidth]{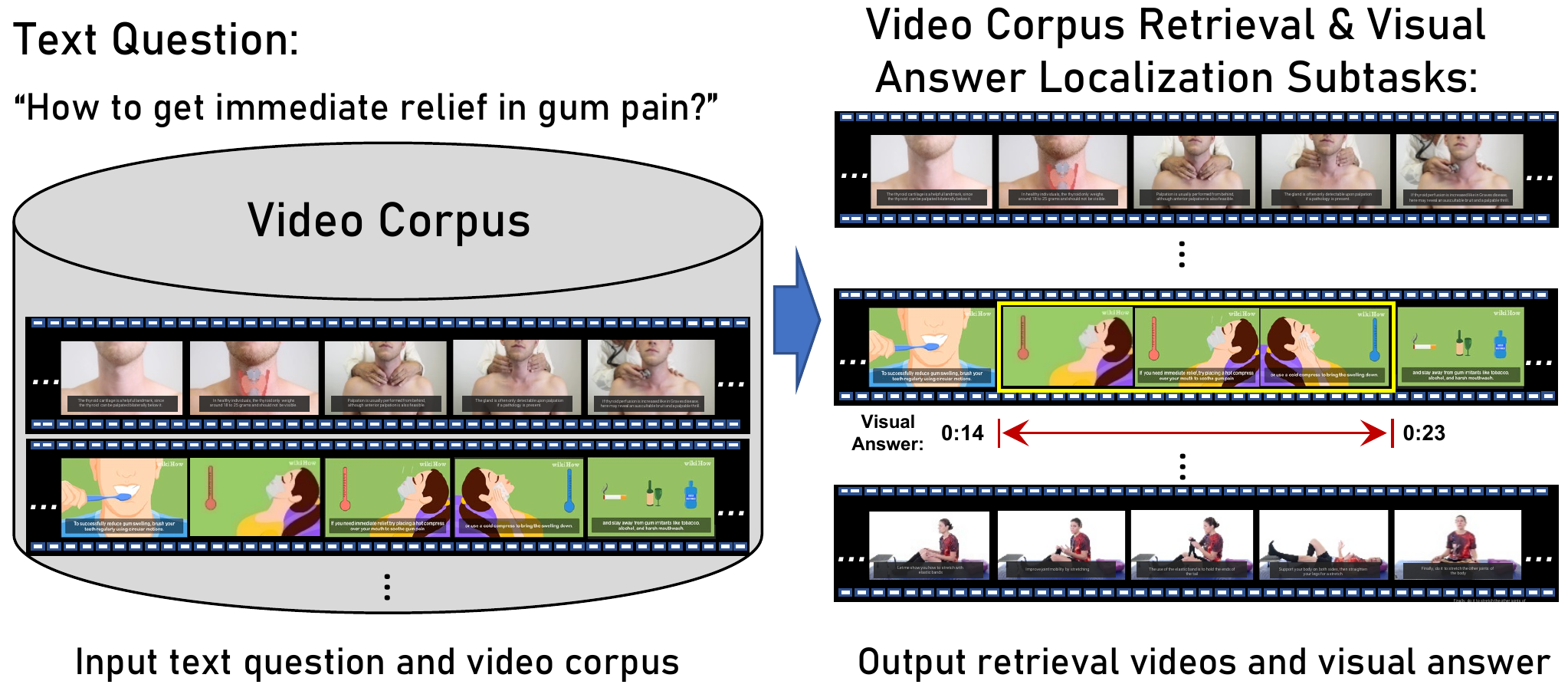}
		\vspace{-0.1cm}
		\caption{Illustration of the video corpus visual answer localization in the medical instructional video, where the visual answer with the subtitles is highlighted in the yellow box.
		}
				\vspace{-0.4cm}
		\label{sample2}
	\end{figure}

	Medical video question answering (MedVidQA) \cite{gupta2022dataset} is the pioneer for the VAL task within one untrimmed video, which employs medical experts to annotate the corpus manually. However, it orients towards a single video and presumes the visual answer exists in the given video, which greatly limits the human-machine interaction application scenarios in the vast videos \cite{chai2021deep, zhang2021video}. Therefore, we extend and introduce a new video corpus visual answer localization (VCVAL) task shown in Fig.~\ref{sample2}. Specifically, we reconstruct the MedVidQA datasets into the medical video corpus question answering (MedVidCQA) dataset to perform the VCVAL task. Our goal for the VCVAL is to find the matching video answer span with the corresponding question from the large-scale video corpus.{{
	The VCVAL task is more challenging than the original VAL task because it not only requires retrieving the target video from a large-scale video corpus, but also needs to locate the visual answer corresponding to the target video accurately.}}
	This task requires a range of skills—interaction between vision and language, video retrieval, passage comprehension, and visual answer localization. To complete the VCVAL task, two types of problems need to be considered. The first is that the features are inconsistent in cross-video retrieval, where retrieval performance will be worse due to the different features between video contents and the given question \cite{dzabraev2021mdmmt}. The second {{are}} the semantic gaps in the cross-modal modeling between the given retrieved video and the question before visual answer localization, resulting in worse downstream performance \cite{lan2021survey}. 
	\par
	\begin{figure*}[t]
	\centering
	\includegraphics[width=16.6cm]{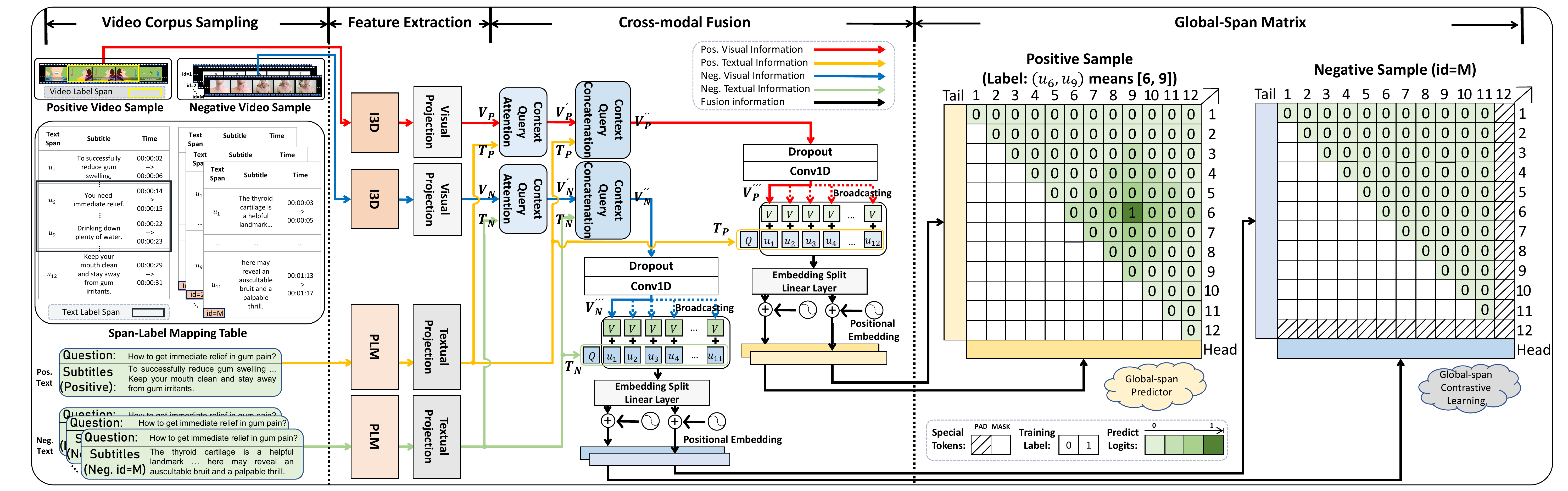}
	\vspace{-0.1cm}
	\caption{Overview of the proposed cross-modal contrastive global-span (CCGS) method, where the modules in the same color represent they share the same parameters.}
	\label{framework}
	\vspace{-0.7cm}
\end{figure*}
	To solve the aforementioned two problems, we propose a cross-modal contrastive global-span (CCGS) method, jointly training the video retrieval and visual answer localization subtasks in an end2end manner. Specifically, to alleviate the retrieval errors caused by feature inconsistency, we design the global-span contrastive learning with the global-span matrix, which sorts the positive and negative {{span points}} to find the correct video.
	Then we enhance the video question-answering semantic by element-wise cross-modal fusion. Finally, we propose a global-span predictor to locate the visual answer, which models joint semantic space with fusion information.
	\par
	In this work, our contributions are presented as follows:
	1) we extend the MedVidQA into the MedVidCQA dataset and introduce the new video corpus visual answer localization (VCVAL) task, which extends the downstream human-machine interaction application scenarios in the vast videos. 2) We propose a novel cross-modal contrastive global-span (CCGS) method. Specifically,  we design the global-span contrastive learning to sort the positive and negative span points in the video corpus. 3) We design the global-span predictor with the element-wise cross-modal fusion to locate the final visual answer. Extensive experiments are carried out for the VCVAL task to demonstrate the effectiveness of the proposed method. The results show that our method outperforms other competitive baselines in both video retrieval and visual answer localization subtasks.
	\vspace{-0.45cm}
	\section{Method}\label{section1: selfmodel}
	\label{sec: Intro}
	\vspace{-0.25cm}
		We propose a cross-modal contrastive global-span (CCGS) method for the VCVAL task, where the overview of the method is presented in Fig.~\ref{framework}. {{The proposed method}} contains four parts: (1) video corpus sampling: to construct the positive and negative samples for contrastive training. (2) Feature extraction: the extracted visual and textual features are processed through the cross-modal interaction. (3) Cross-modal fusion: the visual information {{is added}} to the textual information element-wisely, where the position embedding is also adopted. (4) Global-span matrix: the fusion information is split into two features with position embedding, which is used for constructing the global-span matrix. Two modules adopting the global-span matrix include global-span contrastive learning and global-span predictor.
		\par
		\noindent\textbf{Video Corpus Sampling.}
		\label{sec:cross}
		Given a set of untrimmed videos as $V_C=\{V_i\}_{i=1}^{n}$, corresponding subtitles $S_C=\{S_i\}_{i=1}^{n}$, $S_i = [u_1, \ldots, u_k]$ and the text question as $Q = \{q_j\}^{p}_{j=1}$, our goal is to locate the visual answer in a large corpus, where $S_i$ is each subtitle, $u_k$ is the {{text}} span within the $i$-th subtitle, $k$ is the span number of the corresponding subtitle, $n$ is number of videos in the corpus, and $p$ is length of question tokens. We select the positive and negative samples for contrastive learning, where the videos and texts are processed separately. Also, we construct the span-label mapping table for bridging the visual time with the text span. For convenience, we take the positive sample and the {M}-th negative sample in a batch as an example. As shown in Fig.~\ref{framework}, the ground-truth visual time 00:00:14 $ \sim $ 00:00:23 is corresponded to the $u_6$ $ \sim $  $u_9$.
		\par
		\noindent\textbf{Feature Extraction.}
		For each video $V_i$, we extract visual frames (16 frames per second) to obtain the corresponding RGB visual features $\mathbf{V}_i=\{\mathbf{V}_i^{j}\}_{j=1}^{m}  \in \mathbb{R}^{m \times d}$ using 3D ConvNet~ (I3D) pre-trained on the Kinetics dataset \cite{carreira2017quo}, where $m$ is the number of extracted features and $d$ is the dimension of feature extraction model. The $\mathbf{V}_\text{P}$ and $\mathbf{V}_\text{N}$ through visual projection are represented as the information of the positive (Pos.) and negative (Neg.) samples, where \{$\mathbf {V}_\text{P}$,  $\mathbf {V}_\text{N}$\} $\in \{\mathbf {{V}}_{i}\}^{n}_{i=1}$.
		For the text part, the input question is concatenated with the subtitle of each video $S_i$ to form positive and negative samples. The textual feature representation $\mathbf {{T}}_{i} = \text{PLM}[Q,S_i]$ can be obtained from the pre-trained language model (PLM), where $\mathbf {{T}}_{i}\in\mathbb{R}^{(p + r) \times d}, i\in [1, n]$ and $r$ is the token length of each subtitle. These Pos. and Neg. samples are $\mathbf {T}_\text{P}$ and $\mathbf {T}_\text{N}$, which are chosen from subtitles, where \{$\mathbf {T}_\text{P}$,  $\mathbf {T}_\text{N}$\} $\in \{\mathbf {{T}}_{i}\}^{n}_{i=1}$. 
		\par
		\noindent\textbf{Cross-modal Fusion.}
		Each pair of positive and negative samples are sent into the Context Query Attention \cite{zhang2020span} separately, which leverages the cross-modal modeling through context-to-query  ($\mathcal{A}$) and query-to-context ($\mathcal{B}$) processes. We select question part $\mathbf{{T}}_i^Q \in \mathbb{R}^{p\times d}$ from corresponding $\mathbf{{T}}_i$ as input. \begin{equation}
			\mathcal{A}_i=\{\mathcal{S}_{r}\cdot\mathbf{{T}}_i^Q\}\in\mathbb{R}^{m\times d}, \mathcal{B}_i=\{\mathcal{S}_{c}\cdot\mathcal{S}_{r}^{T}\cdot\mathbf{{V}}_i\}\in\mathbb{R}^{m\times d}\nonumber
		\end{equation}
		where $\mathcal{S}_{r} \in \mathbb{R}^{m \times p}$ and $\mathcal{S}_{c} \in \mathbb{R}^{m \times p}$ are the row- and column-wise normalization of $\mathcal{S}$ by SoftMax, respectively. The output $\mathbf{ V^{\prime}}_i \in \mathbb{R}^{m \times d}$ is written as follows,
		\begin{equation}
			\mathbf{ V^{\prime}}_i=\text{FFN}\big([\mathbf{{V}}_i;\mathcal{A}_i;\mathbf{{V}}_i\odot\mathcal{A}_i;\mathbf{{V}}_i\odot\mathcal{B}_i]\big)
		\end{equation}
		where the $\text{FFN}$ is one layer perceptron network, and $\odot$ denotes element-wise multiplication.
		\par The Context Query Concatenation module \cite{li2022towards} is adopted     to capture deeper semantics from the query with Attention mechanism \cite{chorowski2015attention} and Conv1d (in\_channels = $2d$, out\_channels = $d$), which is calculated as follows, where $\mathbf{ V^{\prime\prime}}_i \in \mathbb{R}^{m \times d}$.
		\begin{equation}
			\mathbf{ V^{\prime\prime}}_i={\text{Conv1d}}\big\{\text{Concat}[\text{Attention}(\mathbf{{V^{\prime}}}_i,\mathbf{T}^Q_i);\mathbf{T}^Q_i]\big\}
		\end{equation}
		
		\par
		The visual information $\mathbf{ V^{\prime\prime\prime}}_i \in \mathbb{R}^{1 \times d}$ is processed with Dropout (p = 0.1) and Conv1d (in\_channels = $d$, out\_channels = $1$) shown as follows.
		\vspace{-0.2cm}
		\begin{equation}
			\mathbf{ V^{\prime\prime\prime}}_i={\text{Conv1d}}\{\text{Dropout}(\mathbf{V^{\prime\prime}}_i)\}
		\end{equation}
		\par
		To obtain well cross-modal fusion information $\overline{\mathbf T_i}$, we select the subtitle spans part ${\mathbf T_i}[p:] \in \mathbb{R}^{ r \times d}$ without the {{question}} interval, and add the cross-modal visual information $\mathbf{V^{\prime\prime\prime}}_i$ element-wisely through the broadcast mechanism. 
		\begin{equation}
			\overline{\mathbf T_i} = \mathbf{V^{\prime\prime\prime}}_i + \mathbf{T}_i[p:]
		\end{equation}
		\par
		Then, the fusion information $\overline{\mathbf T_i}$ is sent into the Embedded Split Linear Layer (ESLayer), where input and output dimension is $[d, {2\times d}]$. Once splitting the last dimension $d$, we can obtain the X-axis feature ($\mathcal{X}$) and Y-axis features ($\mathcal{Y}$) to form the global-span matrix, $\{\mathcal{X}$, $\mathcal{Y}\} \in \mathbb{R}^{r \times d}$, where two features can be {presented} as follows.
		\begin{equation}
			{\mathcal{X}=\text{ESLayer}(\overline{\mathbf T_i})[:d]},~ {\mathcal{Y}=\text{ESLayer}(\overline{\mathbf T_i})[d:]}
		\end{equation}
		\par
		\noindent\textbf{Global-span Matrix.}
		We construct the global-span matrix for jointly modeling the video retrieval and visual answer localization. To better percept different span points in the matrix, we adopt the spiral position embedding\footnote{https://kexue.fm/archives/8265} before the matrix construction, where $P = \{p_j\}_{j=1}^r$. Features $\mathcal{X}$  and $\mathcal{Y}$ are respectively added with spiral linear embedding,
		\vspace{-0.1cm} 
		\begin{equation}
			\hat{\mathcal{X}} = \mathcal{X} + P, ~
			\hat{\mathcal{Y}} = \mathcal{Y} + P	
		\end{equation}
		\par
		\vspace{-0.1cm}
		{The $\hat{\mathcal{X}} $ and $\hat{\mathcal{Y}}$ are processed with dot product for global-span matrix ($\mathbf{Matrix}$)}, where each span point is the unique interval position.
		\begin{equation}
			\mathbf{Matrix} = \text{Average\_Pooling}(\hat{\mathcal{X}} \cdot \hat{\mathcal{Y}}^\top) \in \mathbb{R}^{r \times r}
		\end{equation}
		\par
		\noindent\textbf{Global-span Predictor.} Taking Y-axis as head and X-axis as the tail, we define the span point $[\mathbf{y}, \mathbf{x}]$ to represent the target {visual answer span interval}, where $\mathbf{y}\leq \mathbf{x}$. As shown in the Fig.~\ref{framework}, the label span ($u_6, u_9$) represents the span point [9, 6]. Meanwhile, we mask the region of {$\mathbf{y} ~\textgreater ~\mathbf{x}$} and flatten the $\mathbf{Matrix} \in \mathbb{R}^{1 \times (r \times r)}$ to get local $\mathbf{Matrix}_{\textbf{L}}^{\prime}$ for next process.
		\begin{equation}
			\mathbf{Matrix}_{\textbf{L}}^{\prime} = \text{Flatten}(\mathbf{Matrix}) 
			\label{fla}
		\end{equation}
		\par
		We adopt Cross-Entropy (CE) loss function to maximize the logits of the target span point $[\mathbf{y}, \mathbf{x}]$. And, we convert the point $[\mathbf{y}, \mathbf{x}]$ to the flattened point for further calculation.
		\begin{equation}
			\mathbf{Loss_1} = \mathrm{CE}( \mathbf{Matrix}_{\textbf{L}}^\prime, \text{OneHot}[\mathbf{y} \times d + \mathbf{x}])
		\end{equation}
		\par
		\noindent\textbf{Global-span Contrastive Learning.} To jointly model the semantics in cross-video retrieval, we introduce positive-negative sample pairs with the global-span matrix. As shown in Fig.~\ref{framework}, the exceed area is padded, and all the label is 0.
		We adopt a ratio of positive and negative samples of 1:M, where these samples go through the {flattening} process shown in equation (\ref{fla}) to obtain the output $\mathbf{Matrix_{P}^{\prime}}$, $\{\mathbf{{Matrix_{Ni}^{\prime}}\}^{\text M}_{i=1}}$ respectively. When setting the label, we set the ground truth span point in $\mathbf{Matrix_{P}}$ as the positive {label 1}, the surplus as the negative {label 0}, and each span point in $\{\mathbf{{Matrix_{Ni}^{\prime}}\}^{\text M}_{i=1}}$ is the negative {label 0}.
		%
		\par
				\begin{table*}[t]
			\centering
			\caption{Performance comparisons between the various baseline methods on the MedVidCQA dataset, where four types of baseline methods are compared including visual-based, textual-based, cross-modal, and pipeline-based methods.}
						\vspace{-0.2cm}
			\resizebox{1\textwidth}{!}{
				\begin{tabular}{cc|cccc|cccc|cccc}
						\noalign{\hrule height 1pt}
					\multicolumn{2}{c|}{\thead{\multirow{2}{*}{\begin{tabular}[c]{@{}c@{}} \normalsize {Method} \end{tabular}}}}&\multicolumn{4}{c|}{Rank@1}&\multicolumn{4}{c|}{Rank@10}&\multicolumn{4}{c}{Rank@100}    \\ 
					\multicolumn{2}{c|}{}& IoU=0.3&IoU=0.5&IoU=0.7&mIoU& IoU=0.3&IoU=0.5&IoU=0.7&mIoU& IoU=0.3&IoU=0.5&IoU=0.7&mIoU \\ \hline
					\thead{\multirow{2}{*}{\begin{tabular}[c]{@{}c@{}} \normalsize {Visual-based}\end{tabular}}}&VSLNet \cite{zhang2020span}&2.76&2.07&1.38&1.95&27.27&24.24&12.12&23.20&30.34&22.76&13.10&24.18 \\
					&ACRM \cite{tang2021frame}&3.45&2.07&0.69&1.68&12.12&9.09&9.09&8.74&14.48&10.34&6.21&12.29\\
					
					\hline
					\thead{\multirow{1}{*}{\begin{tabular}[c]{@{}c@{}} \normalsize {Textual-based}\end{tabular}}}&Span-Base \cite{li2022towards}&9.66&6.90&4.14&5.64&38.45&35.15&26.15&30.45&50.68&46.37&37.45&42.68 \\
					\hline
					\thead{\multirow{1}{*}{\begin{tabular}[c]{@{}c@{}} \normalsize {Cross-Modal}\end{tabular}}}& VPTSL \cite{li2022towards}&11.03&10.34&8.28&9.05&39.89&36.48&27.66&32.10&54.32&50.09&40.54&48.87\\

					\hline\hline
					\thead{\multirow{4}{*}{\begin{tabular}[c]{@{}c@{}} \normalsize {BM25 \cite{robertson2004simple}}\end{tabular}}}&VSLNet \cite{zhang2020span}&22.07&13.79&10.34&17.45&37.67&31.55&24.67&26.45&56.12&47.11&38.45&41.69 \\
					&ACRM \cite{tang2021frame}&13.10&8.97&3.45&9.89&24.45&21.43&14.45&20.64&39.96&34.17&30.78&32.89 \\
					&Span-Base \cite{li2022towards}&35.45&28.46&18.44&28.64&53.13&48.66&40.97&47.83&68.97&61.81&49.44&56.84\\
					&VPTSL \cite{li2022towards}&36.56&29.16&17.89&28.89&55.34&49.36&41.23&46.45&67.91&62.41&53.48&59.87\\ \hline
					\thead{\multirow{4}{*}{\begin{tabular}[c]{@{}c@{}} \normalsize {DPR} \cite{karpukhin2020dense}\end{tabular}}}&VSLNet \cite{zhang2020span}& 23.87&14.66&11.08&17.02&38.92&32.56&26.45&27.69&57.89&48.23&40.11&40.98\\
					&ACRM \cite{tang2021frame}&14.48&10.34&5.52&11.70&22.59&19.78&12.15&15.67&56.54&48.65&39.68&40.57\\
					&Span-Base \cite{li2022towards}&38.68&31.78&23.45&30.48&56.73&48.45&43.45&48.35&70.01&63.45&52.63&59.73\\
					&VPTSL \cite{li2022towards}&47.59&38.62&31.03&37.94&63.12&57.15&51.68&58.12&74.86&67.87&60.37&66.91\\ \hline
					\multicolumn{2}{c|}{Our Method}&\textbf{59.31}&\textbf {48.28}&\textbf {38.62}&\textbf {47.00}&\textbf {80.17}&\textbf {70.34}&\textbf {58.62}&\textbf {66.09}&\textbf {91.72}&\textbf {88.28}&\textbf {81.38}&\textbf {82.16} \\ 
						\noalign{\hrule height 1pt}

			\end{tabular}}
			\vspace{-0.2cm}

			\label{t1}
		\end{table*}
%
%
%
\begin{table*}[t]
	\centering
	\caption{Ablation experiments of the proposed method, where CF means Cross-modal Fusion and GS means Global-Span.}
	\vspace{-0.2cm}
	\resizebox{1\textwidth}{!}{
		\begin{tabular}{c|cccc|cccc|cccc|cccc}
			\noalign{\hrule height 1pt}
			\multicolumn{1}{c|}{\thead{\multirow{2}{*}{\begin{tabular}[c]{@{}c@{}} \normalsize {Method} \end{tabular}}}}&\multicolumn{4}{c|}{Rank@1}&\multicolumn{4}{c|}{Rank@10}&\multicolumn{4}{c|}{Rank@100}&\multicolumn{4}{c}{Retrieval}    \\ 
			%
			%
			
			& IoU=0.3&IoU=0.5&IoU=0.7&mIoU& IoU=0.3&IoU=0.5&IoU=0.7&mIoU& IoU=0.3&IoU=0.5&IoU=0.7&mIoU&MRR&R@1&R@5&R@10 \\ \hline
			
 			Our Method
&\textbf{59.31}&\textbf {48.28}&\textbf {38.62}&\textbf {47.00}&\textbf {80.17}&\textbf {70.34}&\textbf {58.62}&\textbf {66.09}&\textbf {91.72}&\textbf {88.28}&\textbf {81.38}&\textbf {82.16} &\textbf{86.70}&\textbf{78.62}&\textbf{97.93}&\textbf{100.00}\\
			\hline
			W/o element-wise CF  &53.10&43.45&29.66&40.94&73.79&66.21&55.86&61.25&86.21&82.07&77.93&78.19&82.44&72.41&94.48&98.31 \\
			W/o GS Contrastive &29.66&24.14&19.31&23.25&45.52&37.93&33.10&37.73&62.76&57.24&50.34&54.59&52.64&40.69&67.59&77.93 \\
			W/o GS Predictor&54.54&45.97&36.45&42.99&68.12&60.29&51.41&57.65&87.48&83.68&78.73&79.68&78.92&72.56&96.54&98.62 \\
			\noalign{\hrule height 1pt}
	\end{tabular}}
	
	 \vspace{-0.2cm}
	
	\label{t2}
	 \vspace{-0.3cm}
\end{table*}

		Finally, we connect $\mathbf{Matrix_{P}^{\prime}}$ with all $\{\mathbf{{Matrix_{Ni}^{\prime}}\}^{\text M}_{i=1}}$ to form the global feature $\mathbf{Matrix_{G}^{\prime}}$.
		\begin{equation}
			\mathbf{Matrix_{G}^{\prime}} = \text{Concat}(\mathbf{Matrix_{P}^{\prime}},\{\mathbf{{Matrix_{Ni}^{\prime}}\}^{\text M}_{i=1}})
		\end{equation}
		\par We adopt the CE loss to calculate the global-span contrastive learning objective.
		\begin{equation}
			\mathbf{Loss_2} = \mathrm{CE}(\mathbf{Matrix_{G}^{\prime}}, \text{OneHot}[\mathbf{y} \times d + \mathbf{x}] )
		\end{equation}
		\noindent\textbf{End2end Training.} We combine the video retrieval and visual answer localization based on global-span matrix, and jointly train the following loss function.
		\begin{equation}
			\mathbf{Loss} = \mathbf{Loss_1} + \mathbf{Loss_2}
		\end{equation}
		\par
		\vspace{-0.6cm}
		\section{Experimental Setting}\label{section: details}
		We reconstruct a dataset {for} the VCVAL task, named MedVidCQA. It can be divided into two subtasks, 1) video retrieval; 2) visual answer localization. Specifically, we keep the distribution and their original labels the same as the MedVidQA \cite{gupta2022dataset}. The number of training, verification, and test sets are 2,710, 145, and 155 respectively.    In the test set, we removed the video ID number corresponding to the problem. As a result, it is required to retrieve the target video ID from the test corpus, and then locate the visual answer. \noindent\textbf{Evaluation Metrics.}  We use 16 evaluation metrics including retrieval and R@1, R@5, R@10 and MRR \cite{ren2021rocketqav2,khattab2020colbert}, and IOU-0.3/0.5/0.7/mIoU in Rank@1/10/100 \cite{li2022towards, zhang2021video}, to evaluate the performance of different methods on the VCVAL task. We select strong baselines of VAL with uni-modal and cross-modal to benchmark the performance on the VCVAL task. In order to ensure fair {comparisons}, all the compared methods follow the \ parameter settings of the original work.
		\noindent\textbf{Hyperparameters of the Baselines.}
		We use the same pre-trained language model\footnote{{https://huggingface.co/microsoft/deberta-v3-base}} to encode the texts, and then calculate the logits for all videos. {We sort all the logits and select the highest prediction interval as the final result.} We set $d$ = 768 and optimize the loss function via the AdamW optimizer with $lr$ = 1e-5. 
		\noindent\textbf{Hyperparameters of the CCGS.
		}
		The hidden size $d$ is set to 768. We use DeBERTa-v3-base as the PLM (with limited maximum length to 1300). We conduct experiments with the Pytorch \cite{paszke2019pytorch} on three NVIDIA A100 GPUs. Our method is optimized with AdamW optimizer ($lr$ = 1e-5). The M is set to 1, and the batch\_size = 2.
		\par We repeated the experiment three times to reduce the random errors, and selected the best model in the validation set to test the effect in the test set for all experiments.
		\begin{table}[t]
	\centering
	\caption{Retrieval performance of different baselines.}
	\vspace{-0.2cm}
	\resizebox{0.485\textwidth}{!}{
		\setlength{\tabcolsep}{5.0mm}
		\begin{tabular}{c|cccc}
			\noalign{\hrule height 1pt}
			\multicolumn{1}{c|}{\thead{\multirow{2}{*}{\begin{tabular}[c]{@{}c@{}} \normalsize {Method} \end{tabular}}}}&\multicolumn{4}{c}{Retrieval}    \\ 
			\multicolumn{1}{c|}{}& MRR&R@1&R@5&R@10 \\ \hline
			VSLNet \cite{zhang2020span}&1.64&2.76&14.55&22.76 \\
			ACRM \cite{tang2021frame}&1.12&3.45&11.21&22.76 \\
			
			
			Span-Base \cite{li2022towards}&10.45&9.66&34.56&41.65 \\
			
			VPTSL \cite{li2022towards}&15.64&11.70&42.56&54.12\\
			
			
			BM25 \cite{robertson2004simple}&70.71&64.83&85.63&94.17 \\
			
			DPR \cite{karpukhin2020dense} &74.56&68.28&89.58&96.54\\ \noalign{\hrule height 0.5pt}
			
			Our Method&\textbf{86.70}&\textbf{78.62}&\textbf{97.93}&\textbf{100.00} \\ 
			\noalign{\hrule height 1pt}
	\end{tabular}}
	 \vspace{-0.6cm}
	\label{t3}
\end{table}
		\vspace{-0.27cm}
		\section{PERFORMANCE EVALUATION}
		As shown in Table \ref{t1}, we have benchmarked the VCVAL task from four perspectives, i.e., visual-based, textual-based (only use subtitle), cross-modal, and pipeline-based methods (three types of methods combined with BM25 or DPR). Our method has achieved impressive results on all metrics, as it builds a unified global-span matrix {among} different videos. {And, each logits in the matrix represents the probability distribution of its answer interval.} Compared with the former three compared methods, our method models the beginning and ending span points jointly, which performs better in video retrieval and answer localization. 
		Table \ref{t3} shows the retrieval performance, where the text-passage retrieval methods are strong baselines as they model the semantics between the textual question and answer. We propose global-span contrastive learning utilizing the global-span matrix, which jointly models semantics in different videos. Meanwhile, it also outperforms the pipeline-based methods shown in Table \ref{t1}. The reason may be that the visual answer localization performance is limited by the retrieval performance due to the pipeline form. \par
		\noindent\textbf{Ablation Studies.}
		Table \ref{t2} presents the ablation experiments. In the video retrieval subtask, global-span contrastive learning contributes the most, as it increases the consistency of video retrieval features using the global-span matrix. These enhanced semantic features have a positive benefit on the performance of video corpus retrieval. In addition, by deleting the element-wise cross-modal fusion (only using subtitles) and the global-span predictor, the performances of our method show varying degrees of degradation. These demonstrate the effectiveness of the proposed method for the VCVAL task.
		\par
		\vspace{-0.4cm}
		\section{Conclusion}
			\vspace{-0.1cm}
		In this paper, we reconstructed a dataset MedVidCQA and introduced a new VCVAL task, which aims to locate the visual answer in a large collection of instructional videos with the text question. We designed the global-span predictor and global-span contrastive learning based on the element-wise cross-modal fusion, modelling the VCVAL task in an end2end manner. Our method outperformed the competitive baseline methods including visual-based, textual-based, cross-modal and pipeline-based methods in terms of video retrieval and answer localization. Besides, the ablation study demonstrated the effectiveness of the proposed method. We hope our work can bring more thoughts on video applications. In the future, we plan to build more instructional video datasets to advance the development of related fields.
		\par
		

		\clearpage
		\bibliographystyle{IEEEbib}
		\bibliography{strings,refs}

\begin{thebibliography}{10}

\bibitem{koupaee2018wikihow}
Mahnaz Koupaee and William~Yang Wang,
\newblock ``Wikihow: A large scale text summarization dataset,''
\newblock {\em arXiv preprint arXiv:1810.09305}, 2018.

\bibitem{drozd2018medical}
Brandy Drozd, Emily Couvillon, Andrea Suarez, et~al.,
\newblock ``Medical youtube videos and methods of evaluation: literature
  review,''
\newblock {\em JMIR medical education}, vol. 4, no. 1, pp. e8527, 2018.

\bibitem{gupta2022dataset}
Deepak Gupta, Kush Attal, and Dina Demner-Fushman,
\newblock ``A dataset for medical instructional video classification and
  question answering,''
\newblock {\em arXiv preprint arXiv:2201.12888}, 2022.

\bibitem{li-etal-2022-vpai}
Bin Li, Yixuan Weng, Fei Xia, Bin Sun, and Shutao Li,
\newblock ``{VPAI}{\_}{L}ab at {M}ed{V}id{QA} 2022: A two-stage cross-modal
  fusion method for medical instructional video classification,''
\newblock in {\em Proceedings of the 21st Workshop on Biomedical Language
  Processing}, pp. 212--219.

\bibitem{lei2018tvqa}
Jie Lei, Licheng Yu, Mohit Bansal, and Tamara Berg,
\newblock ``Tvqa: Localized, compositional video question answering,''
\newblock in {\em Proceedings of the 2018 Conference on Empirical Methods in
  Natural Language Processing}, 2018, pp. 1369--1379.

\bibitem{li2022towards}
Bin Li, Yixuan Weng, Bin Sun, and Shutao Li,
\newblock ``Towards visual-prompt temporal answering grounding in medical
  instructional video,''
\newblock {\em arXiv preprint arXiv:2203.06667}, 2022.

\bibitem{gupta2022overview}
Deepak Gupta and Dina Demner-Fushman,
\newblock ``Overview of the medvidqa 2022 shared task on medical video
  question-answering,''
\newblock in {\em Proceedings of the 21st Workshop on Biomedical Language
  Processing}, 2022, pp. 264--274.

\bibitem{chai2021deep}
Junyi Chai, Hao Zeng, Anming Li, and Eric~WT Ngai,
\newblock ``Deep learning in computer vision: A critical review of emerging
  techniques and application scenarios,''
\newblock {\em Machine Learning with Applications}, vol. 6, pp. 100134, 2021.

\bibitem{zhang2021video}
Hao Zhang, Aixin Sun, Wei Jing, Guoshun Nan, Liangli Zhen, Joey~Tianyi Zhou,
  and Rick Siow~Mong Goh,
\newblock ``Video corpus moment retrieval with contrastive learning,''
\newblock in {\em Proceedings of the 44th International ACM SIGIR Conference on
  Research and Development in Information Retrieval}, 2021, pp. 685--695.

\bibitem{dzabraev2021mdmmt}
Maksim Dzabraev, Maksim Kalashnikov, Stepan Komkov, and Aleksandr Petiushko,
\newblock ``Mdmmt: Multidomain multimodal transformer for video retrieval,''
\newblock in {\em Proceedings of the IEEE/CVF Conference on Computer Vision and
  Pattern Recognition}, 2021, pp. 3354--3363.

\bibitem{lan2021survey}
Xiaohan Lan, Yitian Yuan, Xin Wang, Zhi Wang, and Wenwu Zhu,
\newblock ``A survey on temporal sentence grounding in videos,''
\newblock {\em arXiv preprint arXiv:2109.08039}, 2021.

\bibitem{carreira2017quo}
Joao Carreira and Andrew Zisserman,
\newblock ``Quo vadis, action recognition? a new model and the kinetics
  dataset,''
\newblock in {\em proceedings of the IEEE Conference on Computer Vision and
  Pattern Recognition}, 2017, pp. 6299--6308.

\bibitem{zhang2020span}
Hao Zhang, Aixin Sun, Wei Jing, and Joey~Tianyi Zhou,
\newblock ``Span-based localizing network for natural language video
  localization,''
\newblock in {\em Proceedings of the 58th Annual Meeting of the Association for
  Computational Linguistics}, 2020, pp. 6543--6554.

\bibitem{chorowski2015attention}
Jan Chorowski, Dzmitry Bahdanau, Dmitriy Serdyuk, Kyunghyun Cho, and Yoshua
  Bengio,
\newblock ``Attention-based models for speech recognition,''
\newblock in {\em Proceedings of the 28th International Conference on Neural
  Information Processing Systems-Volume 1}, 2015, pp. 577--585.

\bibitem{tang2021frame}
Haoyu Tang, Jihua Zhu, Meng Liu, Zan Gao, and Zhiyong Cheng,
\newblock ``Frame-wise cross-modal matching for video moment retrieval,''
\newblock {\em IEEE Transactions on Multimedia}, vol. 24, pp. 1338--1349, 2021.

\bibitem{robertson2004simple}
Stephen Robertson, Hugo Zaragoza, and Michael Taylor,
\newblock ``Simple bm25 extension to multiple weighted fields,''
\newblock in {\em Proceedings of the thirteenth ACM international conference on
  Information and knowledge management}, 2004, pp. 42--49.

\bibitem{karpukhin2020dense}
Vladimir Karpukhin, Barlas Oguz, Sewon Min, Patrick Lewis, Ledell Wu, Sergey
  Edunov, Danqi Chen, and Wen-tau Yih,
\newblock ``Dense passage retrieval for open-domain question answering,''
\newblock in {\em Proceedings of the 2020 Conference on Empirical Methods in
  Natural Language Processing (EMNLP)}, 2020, pp. 6769--6781.

\bibitem{ren2021rocketqav2}
Ruiyang Ren, Yingqi Qu, Jing Liu, Wayne~Xin Zhao, Qiaoqiao She, Hua Wu, Haifeng
  Wang, and Ji-Rong Wen,
\newblock ``Rocketqav2: A joint training method for dense passage retrieval and
  passage re-ranking,''
\newblock {\em arXiv preprint arXiv:2110.07367}, 2021.

\bibitem{khattab2020colbert}
Omar Khattab and Matei Zaharia,
\newblock ``Colbert: Efficient and effective passage search via contextualized
  late interaction over bert,''
\newblock in {\em Proceedings of the 43rd International ACM SIGIR conference on
  research and development in Information Retrieval}, 2020, pp. 39--48.

\bibitem{paszke2019pytorch}
Adam Paszke, Sam Gross, Francisco Massa, Adam Lerer, James Bradbury, Gregory
  Chanan, Trevor Killeen, Zeming Lin, Natalia Gimelshein, Luca Antiga, et~al.,
\newblock ``Pytorch: An imperative style, high-performance deep learning
  library,''
\newblock {\em Advances in neural information processing systems}, vol. 32,
  2019.

\end{thebibliography}
		
	\end{document}